\documentclass[sigconf]{acmart}
\AtBeginDocument{%
  \providecommand\BibTeX{{%
    \normalfont B\kern-0.5em{\scshape i\kern-0.25em b}\kern-0.8em\TeX}}}

\copyrightyear{2022}
\acmYear{2022}
\setcopyright{rightsretained}
\acmConference[FAccT '22]{2022 ACM Conference on Fairness, Accountability, and Transparency}{June 21--24, 2022}{Seoul, Republic of Korea}
\acmBooktitle{2022 ACM Conference on Fairness, Accountability, and Transparency (FAccT '22), June 21--24, 2022, Seoul, Republic of Korea}\acmDOI{10.1145/3531146.3533185}
\acmISBN{978-1-4503-9352-2/22/06}

\usepackage{xcolor,colortbl}
\usepackage{tikz}
\usepackage{pgfplots}
\usetikzlibrary{patterns}
\usepgfplotslibrary{colormaps}
\usetikzlibrary{pgfplots.colormaps} 
\pgfplotscreatecolormap{hot}{color(0cm)=(blue); color(1cm)=(yellow); color(2cm)=(orange); color(3cm)=(red)}

\definecolor{lightgray}{rgb}{0.83, 0.83, 0.83}
\definecolor{darkgray}{rgb}{0.05, 0.05, 0.05}
\definecolor{mediumgray}{rgb}{0.66, 0.66, 0.66}

\usepackage{subfigure}

\begin{document}
\title{Evidence for Hypodescent in Visual Semantic AI}


\author{Robert Wolfe}
\affiliation{%
  \institution{University of Washington}
  \city{Seattle}
  \state{WA}
  \country{USA}}
\email{rwolfe3@uw.edu}

\author{Mahzarin R. Banaji}
\affiliation{%
  \institution{Harvard University}
  \city{Cambridge}
  \state{MA}
  \country{USA}}
\email{mahzarin_banaji@harvard.edu}

\author{Aylin Caliskan}
\affiliation{%
  \institution{University of Washington}
  \city{Seattle}
  \state{WA}
  \country{USA}}
\email{aylin@uw.edu}

\begin{abstract}

We examine the state-of-the-art multimodal "visual semantic" model CLIP ("Contrastive Language Image Pretraining") for the rule of hypodescent, or one-drop rule, whereby multiracial people are more likely to be assigned a racial or ethnic label corresponding to a minority or disadvantaged racial or ethnic group than to the equivalent majority or advantaged group. A face morphing experiment grounded in psychological research demonstrating hypodescent indicates that, at the midway point of $1,000$ series of morphed images, CLIP associates 69.7\% of Black-White female images with a Black text label over a White text label, and similarly prefers Latina (75.8\%) and Asian (89.1\%) text labels at the midway point for Latina-White female and Asian-White female morphs, reflecting hypodescent. Additionally, assessment of the underlying cosine similarities in the model reveals that association with White is correlated with association with "person," with Pearson's $\rho$ as high as $0.82$, $p < 10^{-90}$ over a $21,000$-image morph series, indicating that a White person corresponds to the default representation of a person in CLIP. Finally, we show that the stereotype-congruent pleasantness association of an image correlates with association with the Black text label in CLIP, with Pearson's $\rho = 0.48$, $p < 10^{-90}$ for $21,000$ Black-White multiracial male images, and $\rho = 0.41$, $p < 10^{-90}$ for Black-White multiracial female images. CLIP is trained on English-language text gathered using data collected from an American website (Wikipedia), and our findings demonstrate that CLIP embeds the values of American racial hierarchy, reflecting the implicit and explicit beliefs that are present in human minds. We contextualize these findings within the history of and psychology of hypodescent. Overall, the data suggests that AI supervised using natural language will, unless checked, learn biases that reflect racial hierarchies.
\end{abstract}



\maketitle

\section{Introduction}

Recent progress in multimodal "visual semantic" artificial intelligence (AI) has produced CLIP ("Contrastive Language Image Pretraining"), the first language-and-image model that allows the definition of image classes in natural language and generalizes to datasets on which it was not explicitly trained \cite{radford2021learning}. For the field of computer vision, CLIP was able to realize many of the benefits of large-scale self-supervised pretraining first seen in natural language processing (NLP). While its designers primarily evaluate CLIP in the context of image classification \cite{radford2021learning}, features derived from the model have been used to train zero-shot object detection models and zero-shot image generation AI \cite{gu2021open,ramesh2021zero}. Investigations of biases in CLIP have uncovered under-representation of women in image retrieval tasks \cite{wang2021gender}, the presence of biased and offensive content in an open-source training corpus similar to the one on which CLIP was trained \cite{birhane2021multimodal}, and semantic biases (\textit{e.g.,} association of Muslims with terrorism), in multimodal neuron activations in the CLIP image encoder \cite{goh2021multimodal}. In this research, we investigate whether the manner in which CLIP forms categorical boundaries between images of humans reflects a particular racial bias: the rule of hypodescent, or one-drop rule. 

In the paper introducing CLIP, \citet{radford2021learning} report that in the zero-shot setting, CLIP associates with a White text label only 58.3\% of people with the White racial label in the FairFace dataset \cite{karkkainen2021fairface}, while associating 91.3\% of individuals belonging to the six other racial and ethnic groups represented in FairFace with their FairFace label. While the FairFace dataset consists of visually noisy in-the-wild photographs, and relies not on self-identification of race but on the potentially biased labels of human perceivers (Amazon Mechanical Turkers), the results of \citet{radford2021learning} indicate that a substantial minority of people who are perceived by other humans as White are associated with a race other than White by CLIP. The inverse does not appear to be true, as individuals who are perceived to belong to other racial or ethnic labels in the FairFace dataset are associated with those labels by CLIP. This result suggests the possibility that CLIP has learned the rule of "hypodescent," as described by social scientists: individuals with multiracial ancestry are more likely to be perceived and categorized as belonging to the minority or less advantaged parent group than to the equally legitimate majority or advantaged parent group. In other words, the child of a Black and a White parent is perceived to be more Black than White; and the child of an Asian and a White parent is perceived to be more Asian than White. Psychological research finds that a rule of hypodescent reflects a belief in racial hierarchy, and maintains and enhances the racial status quo \cite{ho2013status}.

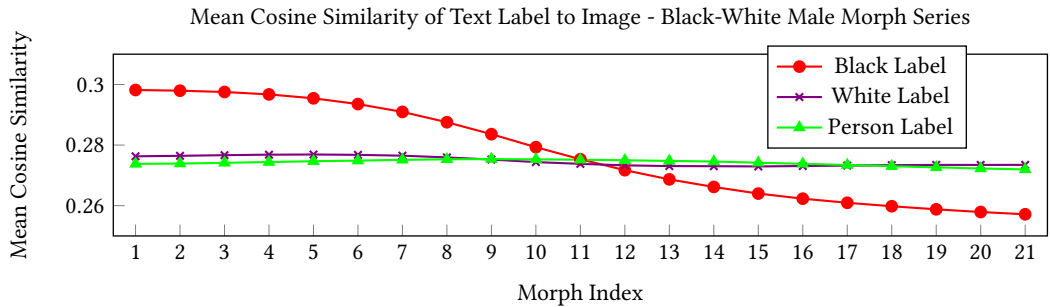
\begin{figure*}[h!]
\begin{tikzpicture}
\begin{axis} [
    height=4cm,
    width=14cm,
    line width = .5pt,
    ymin = 0.25, 
    ymax = 0.31,
    xmin=-.5,
    xmax=20.5,
    ylabel={Mean Cosine Similarity},
    ylabel shift=-5pt,
    xtick = {0,1,2,3,4,5,6,7,8,9,10,11,12,13,14,15,16,17,18,19,20},
    xticklabels = {1,2,3,4,5,6,7,8,9,10,11,12,13,14,15,16,17,18,19,20,21},
    xtick pos=left,
    ytick pos = left,
    title=Mean Cosine Similarity of Text Label to Image - Black-White Male Morph Series,
    xlabel= {Morph Index},
    legend style={at={(.70,.48)},anchor=south west}
]

\addplot[thick,solid,mark=*,color=red] coordinates {(0,0.29818063974380493) (1,0.2979529798030853) (2,0.29752489924430847) (3,0.29672402143478394) (4,0.29544103145599365) (5,0.29353034496307373) (6,0.2909492552280426) (7,0.2875388562679291) (8,0.28359755873680115) (9,0.2793039381504059) (10,0.275326132774353) (11,0.271712988615036) (12,0.2686721980571747) (13,0.2661767899990082) (14,0.2639869153499603) (15,0.2623063623905182) (16,0.2609483003616333) (17,0.25980475544929504) (18,0.2587854266166687) (19,0.2579006850719452) (20,0.257148802280426)};

\addplot[thick,solid,mark=x,color=violet] coordinates {(0,0.2762896716594696) (1,0.27645203471183777) (2,0.27666589617729187) (3,0.2768275737762451) (4,0.2768844962120056) (5,0.2767707407474518) (6,0.27649396657943726) (7,0.27591946721076965) (8,0.2752167284488678) (9,0.27437543869018555) (10,0.27380090951919556) (11,0.2733047902584076) (12,0.2730652987957001) (13,0.2730197608470917) (14,0.27294835448265076) (15,0.2730795741081238) (16,0.27323007583618164) (17,0.2733716368675232) (18,0.27345335483551025) (19,0.273441880941391) (20,0.27346664667129517)};

\addplot[thick,solid,mark=triangle*,color=green] coordinates {(0,0.2737787663936615) (1,0.27389028668403625) (2,0.2741427421569824) (3,0.2743975818157196) (4,0.2746686041355133) (5,0.2748645842075348) (6,0.27513387799263) (7,0.27530035376548767) (8,0.2753828763961792) (9,0.2752758264541626) (10,0.2751733362674713) (11,0.27497294545173645) (12,0.2747740149497986) (13,0.274562269449234) (14,0.2741757035255432) (15,0.2738088071346283) (16,0.27341794967651367) (17,0.27302679419517517) (18,0.2726530134677887) (19,0.2722660005092621) (20,0.27196022868156433)};

\legend {Black Label, White Label, Person Label};

\end{axis}
\end{tikzpicture}
\caption{Across 5\% increments in mixing ratio between Black source images (index 1) and White target images (index 21), the mean probability of a White text label and the mean probability of a person text label are nearly equivalent and nearly invariant. On the other hand, the mean probability of the Black text label varies with the mixing ratio of the images, indicating that White is the default race in CLIP, with other racial and ethnic groups defined based on difference from White.}
\label{mean_cos_sim_bm_wm}
\Description{Figure depicting the mean cosine similarity of for Black, White, and Person labels across the 21 steps of the morph series.}
\end{figure*}

We systematically test for evidence of hypodescent in CLIP. This is, to our knowledge, the first analysis of hypodescent in AI. We make research code public at \url{https://github.com/wolferobert3/evidence_for_hypodescent}. Three main results are highlighted:

\begin{enumerate}
    \item \textbf{CLIP shows hypodescent, or the one-drop rule, by biased association of images of multiracial people with minority ancestry.} Faces of individuals who self-identified as Black, Asian, and Latino/a were morphed into faces of people who self-identified as White. At the halfway point of the morph series, the critical juncture, CLIP associates the majority of the morphed images with a Black, Asian, or Latino/a text label rather than a White text label. Hypodescent is more pronounced for images of women: at the halfway point, 69.7\% of Black-White female multiracial images are associated with Black; 75.8\% of Latina-White female multiracial images are associated with Latina; and 89.1\% of Asian-White female multiracial images are associated with Asian. 53.8\% of Black-White male multiracial images at the halfway point are associated with Black. We do not observe hypodescent for the Asian-White male or for the Latino-White male morph series.
    \item \textbf{White is the default race in CLIP, with other racial and ethnic groups defined by their deviation from White.} For each morph series, we obtain Pearson's $\rho$ for the cosine similarity of a "person" label (with no text indicating race or ethnicity) and the cosine similarity of White, Black, Asian, and Latino/a text labels. In all cases, the correlation between person and White is higher ($\rho \in [0.60,0.82]$) than the correlation between person and Asian, Black, or Latino ($\rho \in [0.11,0.40]$), indicating that the most similar representation to a person is a White person in CLIP. As shown in Figure \ref{mean_cos_sim_bm_wm}, the mean cosine similarity for the White label at each step of the morph series is nearly invariant, and nearly indistinguishable from that of the person label. On the other hand, the mean cosine similarity of the Black label varies as the mixing ratio of the source and target images changes, such that images are associated with Black based on deviation from the White default.
    \item \textbf{Valence bias (association with bad or unpleasant vs. good or pleasant concepts) correlates with association with the minority group.} For each morph series, the SC-WEAT \cite{caliskan2017semantics} is used to measure the association of each embedded image with unpleasant vs. pleasant concepts in the CLIP embedding space, represented using $25$ valenced words each (\textit{e.g.,} grief, agony, disaster vs. love, peace, cheer). The cosine similarity of each image with the minority race or ethnicity text label is obtained. For the $21,000$-image Black-White male morph series, the association with unpleasant concepts for an embedded image correlates with the image's association with the Black label, with Pearson's $\rho = 0.48$, $p < 10^{-90}$. The correlation is similar for the Black-White female morph series, with $\rho = 0.41$, $p < 10^{-90}$. 
\end{enumerate}

\noindent Although this work will discuss racial categories such as Asian, Black, Latino, and White, we use them in the manner in which the social and psychological sciences use them, \textit{i.e.,} to refer to social constructions, not biological realities \cite{american2020new}.

\section{Related Work}

This research draws on multiple strands of previous research on hypodescent, CLIP and multimodal AI, and bias in AI. 

\noindent\textbf{Hypodescent.} Hypodescent refers to the association of people who have multiracial ancestry with the race of their minority parent group. The way multiracial individuals are perceived by others is of sociopolitical importance, as it can serve as an indication of whether intermarriage and miscegenation will result in the disturbance of racial hierarchy\footnote{The term "racial hierarchy" is common in the social and psychological literature, and is the term we will employ here. Others such as \citet{wilkerson2020caste} have characterized the racial ordering of political and economic advantages in America as a caste system. In referring to racial hierarchy, we do not intend to lend legitimacy to this system, but to draw attention to the ways it operates to prevent its encoding in AI.} \cite{ho2011evidence}. For example, if a society has a high number of multiracial births, those at the top of a racial hierarchy may make the hierarchy durable to demographic change by categorizing multiracial individuals as belonging to a minority racial or ethnic group \cite{ho2011evidence}. The minority group increases in size, but can be denied privileges providing access to life's opportunities and outcomes, ranging from housing and loans, to education and jobs, to treatment by law and law enforcement \cite{hickman1996devil}.

The primary psychological reference for the present research is the work of \citet{ho2011evidence}, who uncovered evidence for a rule of hypodescent in human perceivers based on measures of both relatively more implicit and explicit cognition \cite{ho2011evidence}. To study the implicit bias of hypodescent, \citet{ho2011evidence} used a face morphing experiment, wherein the face of a Black person or an Asian person (subjectively strongly judged to be so) is "morphed" across a series of images into a face of a White person. They find that human perceivers are more likely to classify intermediate images according to minority ancestry, reflecting a rule of hypodescent \cite{ho2011evidence}. Moreover, they show that perceivers presented with the family tree of a multiracial person with one Black or Asian parent and one White parent were more likely to classify that person as Black or Asian, respectively, than as White \cite{ho2011evidence}. \citet{ho2011evidence} find that biases of hypodescent are more significant for multiracial people with Black ancestry, but are also observed for multiracial people with Asian ancestry, a result reflecting the existence of American racial hierarchy \cite{hickman1996devil}. As described in the Section \ref{sec:approach}, our research design draws insight from the work of \citet{ho2011evidence} to test for the bias of hypodescent in AI. If the result of \citet{ho2011evidence} is limited to a bias in human minds, it should not be observed in the analyses performed using visual semantic CLIP associations. If, on the other hand, hypodescent is reflected in the CLIP embedding space, the result will add to a growing body of research demonstrating the manner in which human biases are transduced into machine learning architectures.

In a subsequent study of the psychological underpinnings of hypodescent and racial hierarchy, \citet{ho2013status} find that a rule of hypodescent is employed when a minority group makes significant gains in social realms such as business and politics (a threat condition to the racial status quo), and that individuals high in social dominance orientation (an individual difference measure of preference for group-based hierarchies \cite{ho2015nature}) are more likely to categorize multiracial individuals according to a minority parent group. \citet{ho2013status} find that hypodescent can be seen as a "hierarchy-enhancing" social categorization, in that it amplifies perception of racial differences such that racial boundaries and advantages are preserved despite changing demographics and social structures. More recently, \citet{krosch2022threat} find that White perceivers who read about impending demographic shifts which will render the U.S. a majority-minority country are more likely to characterize multiracial faces as belonging to a minority group, reflecting a rule of hypodescent.

Other psychological studies have suggested that hypodescent is related to attention and to frequency of observation. \citet{halberstadt2011barack} suggest that hypodescent is related to selective attention, wherein humans pay more attention to physical characteristics infrequently and first observed later in life, such that people who possess a mixture of majority-group and minority-group traits are perceived as belonging to the minority group. This has particular salience for computer vision, given prior research that finds that people who have darker skin tones, and in particular women who have darker skin tones, are significantly underrepresented in computer vision training datasets \cite{buolamwini2018gender}.  \citet{lewis2016arguing} finds that people who have more experience seeing Black faces are more likely to classify multiracial faces as belonging to a White person. \citet{ho2017you} find that Black perceivers may use hypodescent inclusively, extending Blackness to multiracial people with Black ancestry "because they perceive that Black–White biracials face discrimination and consequently feel a sense of linked fate with them." \citet{peery2008black+} finds that automatic (reflexive/implicit) racial categorizations by White perceivers reflect hypodescent, but that "more complex racial identities may be acknowledged upon more thoughtful reflection." \citet{zarate1990person} find that white perceivers classify members of their own race more quickly, and that speed of categorization is linked to attribution of racial stereotypes.

While hypodescent has been observed for both males and females, \citet{ho2011evidence} observed stronger effects for men than women. This is consistent with research by \citet{sidanius2001social} and \citet{navarrete2010prejudice}, who find that histories of intergroup conflicts involving primarily male aggressors have resulted in stronger exclusionary biases toward outgroup males. The disparity in hypodescent patterns observed in experimental psychology may be a result of broader intersectional gender biases, wherein Black faces are less readily classified as female \cite{goff2008ain}, and male faces are more likely to be classified as Black \cite{carpinella2015gendered}. We are not aware of studies examining hypodescent beyond a gender binary.

\noindent\textit{\textbf{American Context of Hypodescent.}} A notable result that has emerged from the present work is that hypodescent in AI resembles hypodescent in human cognition, as observed by experimental psychologists. Our work focuses on hypodescent in an American context, for two reasons: because most modern psychological research of hypodescent studies American subjects and locates findings within the American context \cite{young2021meta}; and because CLIP is trained on English-language internet data, the majority of which is based on a query list collected from an American website (English-language Wikipedia) \cite{radford2021learning}, suggesting that what CLIP has learned about hypodescent is likely to be best understood within the American context. The historical context of hypodescent in the United States - as a mechanism for expanding slavery \cite{ball2007blurring}, for enforcing segregation \cite{hickman1996devil}, and for preserving the stability of racial hierarchy \cite{liz2018fixity} - has been noted by psychologists as a likely factor in modern Americans' implicit biases with regard to the race of multiracial individuals \cite{ho2011evidence}.

\noindent\textbf{CLIP and Visual Semantic AI.} We now describe the technical architecture of CLIP, the AI model we will test for evidence of hypodescent. CLIP was designed to be the first "zero-shot" image classifier, meaning that it has learned to associate images in computer vision datasets with their text labels without seeing the training data provided for those datasets \cite{radford2021learning}. This constituted a significant step forward in computer vision and multimodal AI, as CLIP advanced the zero-shot state-of-the-art on ImageNet from 11.5\% \cite{li2017learning} to 76.2\% \cite{radford2021learning}. CLIP is known as a "visual semantic" or "multimodal" AI model, in that it encodes text and images and projects them into the same embedding space, such that the text most similar to an image can be matched to it by measuring the cosine similarity of the encoded text and the encoded image \cite{radford2021learning}. While CLIP has been discussed primarily in terms of its performance as a zero-shot image classifier, the model learns "transferable" visual features, meaning that the representations formed by CLIP can be easily adapted for many computer vision tasks, such as image retrieval \cite{wang2021gender}, and can be used to train  specialized zero-shot computer vision models for tasks such as object detection \cite{gu2021open} and image generation \cite{ramesh2021zero,nichol2021glide}. Moreover, the word and sentence representations formed by CLIP have been shown to be highly semantic, to the point that they set or match state-of-the-art on intrinsic evaluations \cite{wolfe2022contrastive}.

CLIP jointly trains a computer vision model (a ResNet \cite{he2016deep} or a Vision Transformer (ViT) \cite{dosovitskiy2020image}) and a contextualizing language model (a smaller version of GPT-2 \cite{radford2019language}), and projects representations from each model into the same embedding space. Both GPT-2 and ViT are based on the transformer architecture, a deep learning network which uses self-attention to draw information from anywhere in its input window \cite{vaswani2017attention}. CLIP's success in generalizing across datasets is the result of applying a novel training objective to an internet-scale language-and-image corpus. CLIP trains on the WebImageText (WIT) corpus, a collection of 400 million images and their associated captions scraped from the internet \cite{radford2021learning}. While WIT is not publicly available, \citet{radford2021learning} note that it contains roughly as many words as the text corpus used to train GPT-2. The query list for WIT consists of all words occurring at least 100 times in English language Wikipedia, bigrams from Wikipedia which have high pointwise mutual information, the names of Wikipedia articles, and all WordNet synsets not included in the list \cite{radford2021learning}. CLIP employs a "contrastive learning" pretraining objective, which maximizes the cosine similarity of an encoded caption with its encoded image while minimizing its cosine similarity with all of the other encoded images in the batch \cite{radford2021learning, tian2019contrastive}. When this research refers to an image or text embedding, this means the multimodal representation formed by CLIP after projection to this joint embedding space. Our supplementary materials include a discussion of advances in deep learning visual semantic AI beginning with its origins in 2013.

This research tests for hypodescent in CLIP by comparing the cosine similarities of embedded images and embedded text labels corresponding to four racial or ethnic groups, an approach which reflects the typical zero-shot use of the model during inference \cite{radford2021learning}. Biases observed in the embedding space will be reflected in the many zero-shot settings in which CLIP is used, including image retrieval \cite{wang2021gender}, image classification \cite{radford2021learning}, and other settings which may benefit from transferable visual features. We examine the CLIP-ViT-Base-Patch32 model, which was downloaded from the Transformers library \cite{wolf-etal-2020-transformers} more than one million times during the month prior to this writing alone, accounting for more than 98\% of the downloads of any CLIP model from the library.

\noindent\textbf{Bias in AI.}
Social biases related to gender \cite{bolukbasi2016man}, age \cite{kim2021age}, religion \cite{abid2021persistent}, and sexuality \cite{sheng2019woman} have been observed in AI systems. Our review of the extensive related work in this area focuses on racial bias in AI and on biases observed in CLIP.

\noindent\textit{\textbf{Implicit Bias and AI.}} CLIP offers the first opportunity to study implicit biases in machine-learned visual semantic representations. Where a supervised algorithm is provided with explicit class labels, CLIP is provided, like word embeddings and language models, with text (and accompanying images) collected from the internet. While unsupervised and self-supervised algorithms free machine learning from using a relatively small number of human-curated labels, and allow for the expansion of training datasets without the need for supervised labeling, unsupervised models learn the social biases of the data on which they are trained \cite{caliskan2017semantics,bender2021dangers}. The most direct adaptation of an implicit bias measurement to machine learned representations is the Word Embedding Association Test (WEAT) of \citet{caliskan2017semantics}, which replicated in word embeddings an array of widely shared human associations and social biases previously observed in human subjects in the Implicit Association Test (IAT) \cite{greenwald1998measuring}. The WEAT was expanded to uncover racial, gender, and intersectional biases in contextualized word embeddings formed by language models like BERT and GPT-2 \cite{guo2021detecting,wolfe2022vast}, and in the generative computer vision model Image GPT \cite{steed2021image}. Our research is most similar to these studies, in that we test an AI model for a bias well-known in psychology.

\noindent\textit{\textbf{GANs for Bias Assessment.}} Our study builds on prior work which uses GAN-generated synthetic images to study bias in AI. \citet{denton2019image} use a generative adversarial network (GAN) to generate series of counterfactual images for analyzing unintended biases in facial analysis AI. \citet{li2021discover} use series of images produced by a GAN to discover unknown biases in image classifiers. \citet{ramaswamy2021fair} use a GAN to generate synthetic training data which is more balanced based on protected attributes such as race and gender. Most recently, \citet{lang2021explaining} train a GAN to explain the decisions of an image classifier by discovering the visual attributes which lead to its predictions.

\noindent\textit{\textbf{Racial Bias in Computer Vision.}} \citet{buolamwini2018gender} find that state-of-the-art AI facial recognition systems fail to detect the faces of women with darker skin, and that this problem is traceable to the underrepresentation of women and people with darker skin in widely used facial recognition datasets. \citet{wilson2019predictive} finds that this problem is not limited to facial recognition, and that state-of-the-art object detection systems also perform poorly for people with darker skin. Moreover, emotion detection AI systems have been shown to disproportionately detect unpleasant emotions such as anger in emotionally ambiguous or neutral images of members of minority racial or ethnic groups \cite{rhue2018racial}.\footnote{There are ethical questions related to whether technologies like facial recognition and emotion detection are beneficial, as they are used for purposes of surveillance and social control \cite{leibold2020surveillance}, and recent research focuses on ways to circumvent mass surveillance \cite{guetta2021dodging}. Where such systems are used, the research of scholars such as \citet{buolamwini2018gender} makes clear the over-representation of men and people who have lighter skin in AI training data, and the failure of commercial AI systems to perform equally for women and for people who have darker skin.} \citet{steed2021image} show that Image GPT, which generates image completions given a visual prompt \cite{chen2020generative}, implicitly associates images of Black individuals with images of weapons.

\noindent\textit{\textbf{Racial Bias in NLP.}} \citet{may2019measuring} find evidence of sentence-level racial bias in language models, while \citet{sheng2019woman} show that the text output of language models like GPT-2 contains biases demonstrating low regard for members of marginalized groups, including racial minorities. \citet{wolfe2021low} find that under-representation of women and marginalized racial and ethnic groups in the training corpora of language models produces representations more likely to be biased and overfit to stereotypical pretraining contexts. \citet{dodge2021documenting} show that filtering algorithms used to construct the C4 training corpus remove information by and about people belonging to marginalized populations.

\noindent\textit{\textbf{Bias in CLIP.}} \citet{radford2021learning} and \citet{agarwal2021evaluating} catalogue social biases in CLIP, including that images of people who are labeled as Black in the FairFace dataset \cite{karkkainen2021fairface} are the most likely to be mislabeled as animals. In highlighting multimodal neurons in CLIP which respond to photographic, symbolic, or textual depictions of a concept, \citet{goh2021multimodal} find that the activation of these neurons reflect stereotypes, such as association of terrorism with Muslims. \citet{birhane2021multimodal} find that LAION-400m \cite{schuhmann_2021}, a multimodal image-and-language dataset similar to CLIP's training corpus, contains racial and ethnic slurs and stereotypes, along with images of sexual violence and other obscene adult content.

\section{Data}

We use the Chicago Face Database (CFD), a dataset of images produced for studies of race in psychology \cite{ma2015chicago}. The CFD includes $597$ high-resolution ($2,444$ x $1,718$ pixel) images of male and female study volunteers, with labels (Asian, Black, Latino/a, and White) based on self-identified race or ethnic group \cite{ma2015chicago}. The subjects of the photos faced the camera, and images are standardized such that they are all set against a White background, with the face occupying the same area of the image. CFD subjects were recruited in the United States. The CFD includes photos of all subjects with a "neutral" facial expression, and of a subset with "happy (open mouth)," "happy (closed mouth)," "angry," and "fearful" facial expressions. The experiments of \citet{ho2011evidence} used images of subjects with a neutral facial expression, and for consistency we use only the images with neutral facial expressions for this research.

\noindent\textbf{GAN-Generated Images.} Consistent with the research design of \citet{ho2011evidence}, who produce face morphs between racial or ethnic groups in 5\% increments, we morph faces of people who self-identify as Black, Asian, and Latino/a into faces of people who self-identify as White in the CFD. Specifically, for each morph series, an image of a person who self-identifies as Black, Asian, or Latino/a is selected, and nineteen intermediate images are generated which blend facial features between that image and the image of a person who self-identifies as White. At each morph step, the morphed image becomes less similar to the image of the person who self-identifies as Black, Asian, or Latino/a, and more similar to the person who self-identifies as White. Using a 21-step series allows the definition of 75\%, 50\%, and 25\% mixture ratios between the source and target images at the 6\textsuperscript{th}, 11\textsuperscript{th}, and 16\textsuperscript{th} morph indices respectively. 

\begin{figure}
    \centering
    \includegraphics[width=\columnwidth]{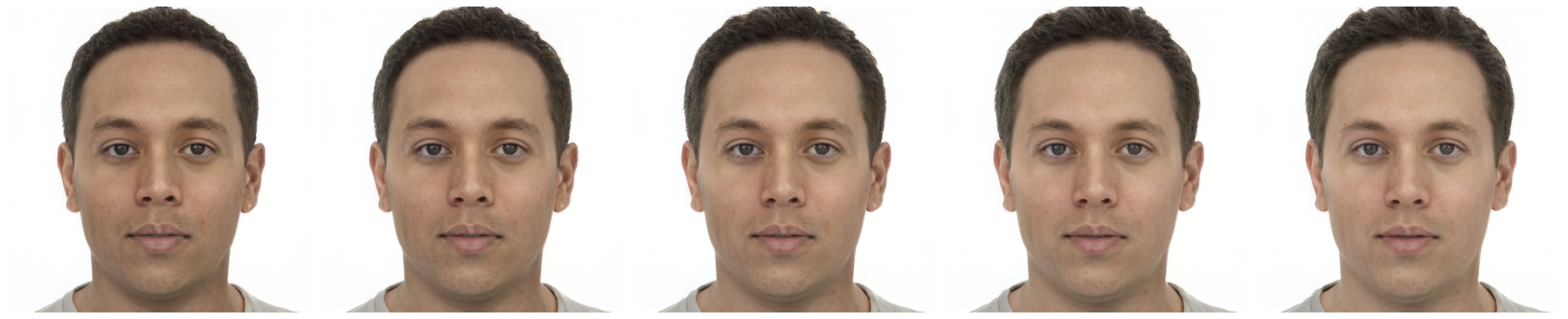}
    \caption{Images of the five middle steps (40-60\% mixing ratio) of a Black-White male GAN-generated morph series. }
    \label{gan_series_example}
    \Description{Figure depicting the middle five images of a GAN-generated series morphing from a self-identified Black individual to a self-identified White individual.}
\end{figure}

For each pair of racial or ethnic groups, we produce $1,000$ unique morph series to ensure the statistical significance of our results. Let $n$ denote the number of images for a "source" racial or ethnic group in the Chicago Face Database, \textit{i.e.,} for the group which serves as the initial image in the morph series, rather than the final image in the morph series. For each image in the source group, we randomly select $\lceil\frac{1,000}{n}\rceil$ (\textit{i.e.,} the largest integer $\geq \frac{1,000}{n}$) images from the target group to serve as final images in the morph series. This method enables maximum variety in the morph series. Because this produces slightly more than $1,000$ morph series per paired source and target group, we randomly downsample to $1,000$ morph series (a total of $21,000$ images) per pair to ensure uniformity between morph series comparisons. To produce highly realistic images, we use StyleGAN2-ADA\footnote{A GAN may itself exhibit bias based on its training dataset; for example, if images of White individuals are overrepresented in the training data, the GAN may lighten the skin tone of generated images, or produce noisy or lower quality images of individuals with physical characteristics underrepresented in the training dataset \cite{jain2020imperfect}. We note that inspection of several thousand generated images by multiple domain experts suggests inter-annotator reliability of these series for quantifying bias.} pretrained on the FFHQ dataset \cite{Karras2020ada}. Images are normalized by cropping around the face, such that facial features appear in positions similar to the positions of the facial features in the pretraining dataset. Failing to do this produces distorted and distended faces bearing little similarity to either of the images between which the morph occurs. To produce morph series using the GAN, we train on the source image and the target image. We use the default hyperparameters of StyleGAN2-ADA, and train for 125 steps per image, beyond which we observe no benefit given the high-resolution, standardized images we provide to the GAN. This produces a source embedding, and a target embedding, for which we use the generator to produce the first and last images in the  series. For the rest of the morph series, we take the difference between these two embeddings and divide it by the number of intermediate images to be produced.\footnote{Steps for morphing between images are derived from the work of \citet{heaton2020applications}.} At each step, we add the divided difference, and create a new image using the generator. This creates a series of $21$ high-resolution (1,024x1,024 pixel) synthetic images, such as those seen in Figure \ref{gan_series_example}. Our supplementary materials include an example of a morph series for each source and target group pair.

A projected image embedding is obtained from CLIP for each of the $21$ images in each morph series. Then, projected text embeddings reflecting the self-identified race or ethnicity of the source group image, the self-identified race or ethnicity of the target group image, three multiracial group labels ("multiracial," "biracial," and "mixed race"), and a "person" label (with race omitted) are obtained from CLIP. Following the prompt of \citet{radford2021learning}, who recommend using "a photo of [[image class]]" for best performance with zero-shot CLIP, we use "a photo of a [[race or ethnicity]] person" as the text input to the model for each of the text labels. 

Note that we do not morph faces across genders. That is, we morph images of men into men, and images of women into women. The reasons for this are three-fold: first, psychological research has tested morphs of men into men and women into women, and we are able to compare our results to prior work in this light.  Second, prior research indicates that the rule of hypodescent is applied more to men than to women, and we intend to test this hypothesis with CLIP. Third, our experiments test for a bias based on race, rather than gender. It is unclear whether gender-related stimuli changing across a series of morphs would introduce noise which distorts the model's association based on race. Future research designs might study how CLIP's biases change when gender stimuli vary along with racial stimuli.

\section{Approach and Experiments}\label{sec:approach}

We conduct three experiments. The first tests whether CLIP associates images of multiracial people with minority ancestry, according to a rule of hypodescent. The second tests whether CLIP embeds White as a default racial group. The third tests a relationship between valence association (good vs. bad) and association with a minority group label.

\noindent\textbf{Test of Hypodescent.} At each step in the morph series, we measure the percentage of morphed images for which the image's cosine similarity with the minority group label is higher than with the White label. If hypodescent is reflected in CLIP, we would expect that the majority of images which blend source features and target features most equally (\textit{i.e.,} those in the middle of the morph series) would be more similar to the embedded text corresponding to the disadvantaged group. We also characterize the skewness of each distribution of associations, defined as $\frac{m_{3}}{m_{2}^{3/2}}$, with the biased $i$th central moment given by $m_{i} = \frac{1}{N} \sum_{1}^{N} (x[n] - \bar{x})^{i}$, and $\bar{x}$ referring to the sample mean. Skewness measures of the symmetry of a distribution: negative skewness indicates that the distribution of associations leans toward the minority racial or ethnic group, while positive skewness indicates a distribution that leans toward White.

\noindent\textbf{Test of White as a Default Race.} For the $21,000$ images of each morph series ($1,000$ images at each morph step), we obtain the cosine similarity with the source (Asian, Black, or Latino/a) group text label, the White text label, and a person label with no race or ethnicity included. We then take Pearson's $\rho$ between the cosine similarities for the person label with those for the minority group label, and with those for the White label. If CLIP encodes White as a default race, we would expect to observe stronger correlations between White and person than between other racial or ethnic groups and person. Next, we take the mean cosine similarity at every morph index (corresponding to a 5\% morph increment) for each racial or ethnic group label with the $1,000$ GAN-generated images at that morph index. Mean cosine similarity can be thought of as the average association with a text class at a morph step. To validate that results using the mean are representative of the full series, we report the standard deviation of $21,000$ cosine similarities for each text label.  If CLIP encodes White as a default race, we would expect to observe correspondence between the cosine similarity for White and person, with other racial and ethnic groups defined in relation to White.

\noindent\textbf{Test of Relationship Between Valence and Race or Ethnicity.} The third analysis focuses on the association of race or ethnicity with valence (good or pleasant vs. bad or unpleasant), which is fundamental to any psychological analysis of social cognition and social interaction \cite{greenwald1998measuring,osgood1964semantic}. Previous research has shown that static word embeddings encode the concept of Black (represented using $25$ African-American names) such that it is more associated with unpleasantness than the concept of White (represented using $25$ European-American names), a result of training on large language corpora scraped from the internet \cite{caliskan2017semantics}. This result confirmed demonstrations of human implicit social cognition showing evidence that even those individuals who report no racial bias nevertheless demonstrate automatic association of Black with bad and White with good \cite{greenwald1998measuring}. The WEAT, and the IAT before it, measure racial bias using two sets of attribute words: one good or pleasant group, and one bad or unpleasant group. These groups correspond to the psycholinguistic property of valence, or the pleasantness or unpleasantness of a stimulus \cite{osgood1964semantic}. Below are the pleasant and unpleasant attribute words used in the IAT and the WEAT, which we also employ:

\noindent \textbf{Pleasant:} caress, freedom, health, love, peace, cheer, friend, heaven, loyal, pleasure, diamond, gentle, honest, lucky, rainbow, diploma, gift, honor, miracle, sunrise, family, happy, laughter, paradise, vacation

\noindent \textbf{Unpleasant:} abuse, crash, filth, murder, sickness, accident, death, grief, poison, stink, assault, disaster, hatred, pollute, tragedy, divorce, jail, poverty, ugly, cancer, kill, rotten, vomit, agony, prison

In addition to measuring the standardized difference in valence between two groups of target words (\textit{e.g.,} European-American names and African-American names), the WEAT is also able to measure the valence of a single target stimulus using the single-category WEAT (SC-WEAT). We adapt the SC-WEAT to measure the valence of a visual semantic representation of an encoded image, rather than a word. The formula for the SC-WEAT is:

\begin{equation}
\frac{\textrm{mean}_{a\in A}\textrm{cos}(\vec{i},\vec{a}) - \textrm{mean}_{b\in B}\textrm{cos}(\vec{i},\vec{b})}{\mathrm{std\_dev}_{x \in A \cup B}\textrm{cos}(\vec{i},\vec{x})}
\end{equation}

\noindent where $\vec{i}$ refers to the multimodal image representation, and $A$ and $B$ refer to multimodal text representations of sets of attribute words. The SC-WEAT returns an effect size, Cohen's $d$ \cite{cohen1992statistical}, and a $p$-value denoting statistical significance. We obtain Pearson's $\rho$ between the SC-WEAT effect size (association with unpleasant vs. pleasant) for the $21,000$ images, with the cosine similarities between the images and a text label corresponding to the minority racial or ethnic group (\textit{i.e.,} the degree of association with the minority group). For this experiment, a positive effect size denotes association with unpleasantness, such that a positive correlation denotes association of a minority racial or ethnic group with unpleasantness, and a negative correlation denotes association with pleasantness.

By relying on visual, face-based stimuli, our research overcomes the limitation imposed by words, as there is little question about whether a face is a face or not, whereas words may contain multiple meanings. Moreover, both research on word embeddings and human subjects research using pictorial stimuli can use only limited inputs, typically $8$ faces or $25$ words, to represent social groups \cite{dasgupta2000automatic}. Using the current methodology expands the stimulus input to be two orders of magnitude greater than the inputs used in previous research with humans or word embeddings. If the present analyses support previous small-scale input studies, the inferences that can be drawn about the strength and generality of group valence associations will be significantly strengthened.

\noindent \textbf{Validating the SC-WEAT for Visual Semantics.} The WEAT and SC-WEAT are well-established methods for measuring biases in linguistic representations \cite{caliskan2017semantics,guo2021detecting,wolfe2021low,wolfe2022vast}, and have been adapted to measure bias in image embeddings by \citet{steed2021image}. As this is the first application of the SC-WEAT to a visual semantic model, we test that using the SC-WEAT produces human-interpretable results. For each image included in the OASIS norms, a dataset of 900 images which includes labels reflecting the human-rated valence of each image \cite{kurdi2017introducing}, we obtain an SC-WEAT effect size measuring the pleasantness of the image's embedding in CLIP. Comparing the SC-WEAT effect sizes with the human-rated valence norms yields Pearson's $\rho=0.77$, $p < 10^{-30}$, on par with similar semantic evaluations of the best static and contextualized word embeddings on valence-labeled linguistic lexica of comparable size \cite{toney2020valnorm,wolfe2022vast}. This means, concretely, that CLIP associates images depicting war, sickness, and so on with unpleasantness; and images related to nature, family, and so on with pleasantness, and that the SC-WEAT can be used to detect these associations.

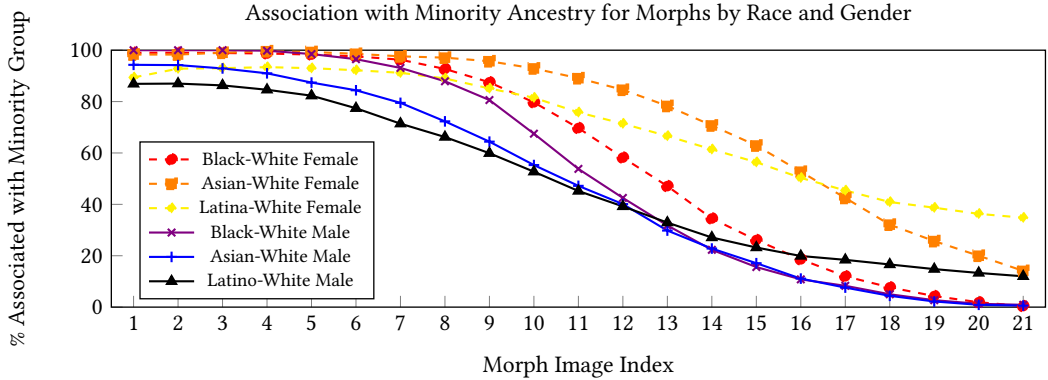
\begin{figure*}[htbp!]
\begin{tikzpicture}
\begin{axis} [
    height=5cm,
    width=14cm,
    line width = .5pt,
    ymin = 0, 
    ymax = 100,
    xmin=-.5,
    xmax=20.5,
    ylabel=\% Associated with Minority Group,
    ylabel shift=-5pt,
    xtick = {0,1,2,3,4,5,6,7,8,9,10,11,12,13,14,15,16,17,18,19,20},
    xticklabels = {1,2,3,4,5,6,7,8,9,10,11,12,13,14,15,16,17,18,19,20,21},
    xtick pos=left,
    ytick pos = left,
    title=Association with Minority Ancestry for Morphs by Race and Gender,
    xlabel= {Morph Image Index},
    legend style={at={(0.03,0.03)},anchor=south west,nodes={scale=.8, transform shape}},
    ]

\addplot[thick,dashed,mark=*,color=red] coordinates {((0,99.0) (1,99.0) (2,98.9) (3,98.7) (4,98.3) (5,97.6) (6,96.3) (7,92.7) (8,87.5) (9,79.7) (10,69.7) (11,58.2) (12,47.1) (13,34.5) (14,26.1) (15,18.6) (16,12.0) (17,7.7) (18,4.3) (19,1.9) (20,0.4)};

\addplot[thick,dashed,mark=square*,color=orange] coordinates {(0,98.3) (1,98.3) (2,99.0) (3,99.6) (4,99.3) (5,98.6) (6,97.6) (7,97.1) (8,95.7) (9,92.9) (10,89.1) (11,84.6) (12,78.3) (13,70.7) (14,62.9) (15,52.7) (16,42.5) (17,32.1) (18,25.7) (19,20.0) (20,14.1)};

\addplot[thick,dashed,mark=diamond*,color=yellow] coordinates {(0,89.4) (1,92.8) (2,93.0) (3,93.4) (4,93.0) (5,92.2) (6,91.2) (7,89.0) (8,85.1) (9,81.5) (10,75.8) (11,71.4) (12,66.6) (13,61.4) (14,56.4) (15,50.3) (16,45.4) (17,40.9) (18,38.7) (19,36.3) (20,34.8)};

\addplot[thick,solid,mark=x,color=violet] coordinates {(0,100.0) (1,100.0) (2,100.0) (3,99.8) (4,98.5) (5,96.5) (6,93.1) (7,87.9) (8,80.6) (9,67.5) (10,53.8) (11,42.5) (12,31.8) (13,22.3) (14,15.6) (15,10.8) (16,8.3) (17,5.0) (18,2.7) (19,1.3) (20,0.9)};

\addplot[thick,solid,mark=+,color=blue] coordinates {(0,94.3) (1,94.2) (2,92.9) (3,91.0) (4,87.4) (5,84.4) (6,79.5) (7,72.3) (8,64.4) (9,55.3) (10,47.2) (11,40.1) (12,29.8) (13,22.8) (14,17.1) (15,11.1) (16,7.6) (17,4.4) (18,2.2) (19,0.9) (20,0.6)};

\addplot[thick,solid,mark=triangle*,color=black] coordinates {(0,86.9) (1,87.0) (2,86.3) (3,84.6) (4,82.3) (5,77.4) (6,71.4) (7,66.2) (8,59.9) (9,52.7) (10,45.2) (11,39.2) (12,32.9) (13,27.1) (14,23.2) (15,19.9) (16,18.4) (17,16.6) (18,14.8) (19,13.3) (20,12.0)};

\legend {Black-White Female, Asian-White Female, Latina-White Female, Black-White Male, Asian-White Male, Latino-White Male};

\end{axis}
\end{tikzpicture}
\caption{Across 5\% increments in mixing ratio between source (minority group) images at index 1 and target (White) images at index 21, CLIP associates a majority of images with the minority group until the image is only 20\% similar to the source group for Asian-White and Latina-White Female morph series, and 40\% similar to the source group for the Black-White Female morph series, indicating evidence of hypodescent for Female images. 53.5\% of Black-White Male images are associated with Black at the 50\% point.}
\label{hypodescent_threshold_all}
\Description{Figure depicting the percentage of the time images are associated with the Asian, Black, or Latino label over the White label across the 21 steps of the morph series.}
\end{figure*}

\section{Results}

Our results provide evidence for hypodescent in the CLIP embedding space, a bias applied more strongly to images of women. Results further indicate that CLIP associates images with racial or ethnic labels based on deviation from White, with White as the default. Finally, an image's valence association correlates with its association with a minority racial label. For readability, some tables below refer to series of Asian-White morphed images as "A-W," of Black-White morphed images as "B-W," and of Latino/a-White morphed images as "L-W."

\begin{table}[htbp]
{
\centering
\begin{tabular}
{|l||r|r|r|r|r|}
 \hline
 \multicolumn{6}{|c|}{Minority Group Associations by Mixing Ratio} \\
 \hline
 Morph Series & 75\% & 55\% & 50\% & 45\% & 25\% \\
 \hline
 Asian-White Female & \cellcolor{gray!99}98.6 & \cellcolor{gray!93}92.9 & \cellcolor{gray!89}89.1 & \cellcolor{gray!85}84.6 & \cellcolor{gray!53}52.7 \\
 Black-White Female & \cellcolor{gray!98}97.6 & \cellcolor{gray!80}79.7 & \cellcolor{gray!70}69.7 & \cellcolor{gray!58}58.2 & \cellcolor{gray!19}18.6 \\
 Latina-White Female & \cellcolor{gray!92}92.2 & \cellcolor{gray!82}81.5 & \cellcolor{gray!76}75.8 & \cellcolor{gray!71}71.4 & \cellcolor{gray!50}50.3 \\
 \hline
 Asian-White Male & \cellcolor{gray!84}84.4 & \cellcolor{gray!55}55.3 & \cellcolor{gray!47}47.2 & \cellcolor{gray!40}40.1 & \cellcolor{gray!11}11.1  \\
 Black-White Male & \cellcolor{gray!97}96.5 & \cellcolor{gray!68}67.5 & \cellcolor{gray!54}53.8 & \cellcolor{gray!43}42.5 & \cellcolor{gray!11}10.8  \\
 Latino-White Male & \cellcolor{gray!77}77.4 & \cellcolor{gray!53}52.7 & \cellcolor{gray!45}45.2 & \cellcolor{gray!39}39.2 & \cellcolor{gray!20}19.9  \\
 \hline
\end{tabular}
}
\caption{At 50\% mixing ratio, the point denoting equal similarity between the source (Asian, Black, or Latino/a) and target (White) images, CLIP associates a higher percentage of $1,000$ morphed Female images with Asian ($89.1\%$), Latina ($75.8\%$), and Black ($69.7\%$) text labels than with a corresponding White label, indicating hypodescent. Hypodescent is also observed for Black-White Male images ($53.8\%$ at 50\% mixing ratio), but not for Asian-White Male or Latino-White Male images.}
\label{intermediate_threshold_table}
\Description{Table displaying the percentage of the time images at five mixing ratios are associated with the Asian, Black, or Latino label over the White label.}
\end{table}

\noindent\textbf{Evidence for hypodescent in CLIP.} For the $1,000$ morph series of $21$ images each between Black males and White males, CLIP assigns a higher cosine similarity to the Black text label $67.5\%$ of the time at a $55\%$ mixing ratio (MR) (\textit{i.e.}, step 9 out of 20, wherein the image is $55\%$ similar to the source (Black) image and $45\%$ similar to the target (White) image), $53.8\%$ of the time at $50\%$ MR, and $42.5\%$ of the time at $45\%$ MR. Hypodescent is not observed at 50\% MR for the Asian-White male or Latino-White male morph series. Female morph series reflect  stronger biases of hypodescent, with $89.1\%$ of Asian-White female morphed images associated with Asian at $50\%$ MR, $69.7\%$ of Black-White female morphed images associated with Black at $50\%$ MR, and $75.8\%$  of Latina-White female morphed images associated with Latina at $50\%$ MR. The Black-White female morph series yields a majority of Black associations until $40\%$ MR, while the Asian-White and Latina-White morph series yield a majority of Asian and Latina associations, respectively, until $20\%$ MR. Full results for 25\%, 45\%, 50\%, 55\%, and 75\% MR steps are provided in Table \ref{intermediate_threshold_table}, and association curves across the six morph series are visualized in Figure \ref{hypodescent_threshold_all}. 

\begin{table}[htbp]
{
\centering
\begin{tabular}
{|l|r|r|r|r|}
\hline
\multicolumn{4}{|c|}{Skew by Morph Series} \\
\hline
Female Series & Skew & Male Series & Skew \\
\hline
A-W Female & \cellcolor{gray!84}-0.84 & A-W Male & 0.00\\
B-W Female & \cellcolor{gray!30}-0.30 & B-W Male & \cellcolor{gray!06}-0.06\\
L-W Female & \cellcolor{gray!42}-0.42 & L-W Male & 0.12 \\
\hline
\end{tabular}
}
\caption{Skew is negative for the distribution of minority group vs. White label associations across all three Female morph series and the Black-White Male morph series. This reflects an asymmetry which leans toward the minority group, indicating hypodescent.}
\label{skew_table}
\Description{Table describing the skew of associations for each morph series from Asian, Black, or Latino to White.}
\end{table}

\noindent Four of the morph series exhibit negative skewness, indicating asymmetry in the distribution of associations which leans toward a racial minority group label. Table \ref{skew_table} shows that results also reflect a stronger bias of hypodescent for women than for men, with a largest male series skew of -0.06 (Black-White male series), and a largest female series skew of -0.84 (Asian-White female series). The smallest female series skew is -0.30 (Black-White female series).

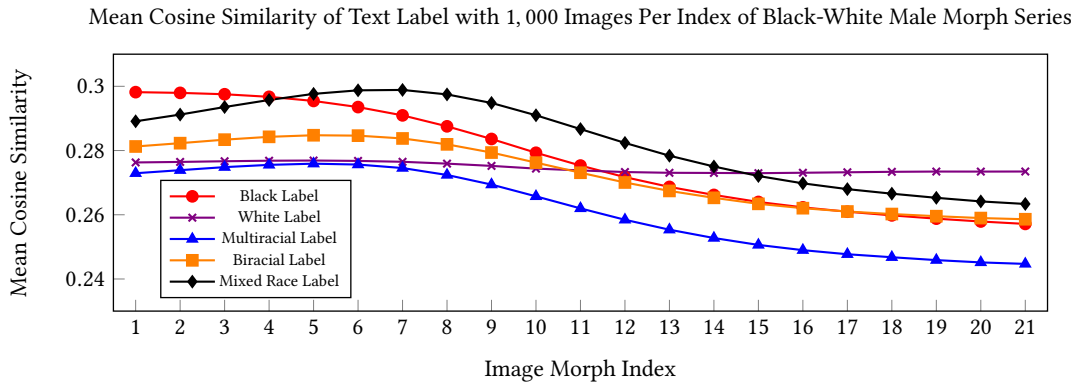
\begin{figure*}[htbp]
\begin{tikzpicture}
\begin{axis} [
    height=5cm,
    width=14cm,
    line width = .5pt,
    ymin = 0.23, 
    ymax = 0.31,
    xmin=-.5,
    xmax=20.5,
    ylabel={Mean Cosine Similarity},
    ylabel shift=-5pt,
    xtick = {0,1,2,3,4,5,6,7,8,9,10,11,12,13,14,15,16,17,18,19,20},
    xticklabels = {1,2,3,4,5,6,7,8,9,10,11,12,13,14,15,16,17,18,19,20,21},
    xtick pos=left,
    ytick pos = left,
    title=Mean Cosine Similarity of Text Label with {$1,000$} Images Per Index of Black-White Male Morph Series,
    xlabel= {Image Morph Index},
    legend style={at={(.05,.05)},anchor=south west,nodes={scale=0.7, transform shape}}
]

\addplot[thick,solid,mark=*,color=red] coordinates {(0,0.29818063974380493) (1,0.2979529798030853) (2,0.29752489924430847) (3,0.29672402143478394) (4,0.29544103145599365) (5,0.29353034496307373) (6,0.2909492552280426) (7,0.2875388562679291) (8,0.28359755873680115) (9,0.2793039381504059) (10,0.275326132774353) (11,0.271712988615036) (12,0.2686721980571747) (13,0.2661767899990082) (14,0.2639869153499603) (15,0.2623063623905182) (16,0.2609483003616333) (17,0.25980475544929504) (18,0.2587854266166687) (19,0.2579006850719452) (20,0.257148802280426)};

\addplot[thick,solid,mark=x,color=violet] coordinates {(0,0.2762896716594696) (1,0.27645203471183777) (2,0.27666589617729187) (3,0.2768275737762451) (4,0.2768844962120056) (5,0.2767707407474518) (6,0.27649396657943726) (7,0.27591946721076965) (8,0.2752167284488678) (9,0.27437543869018555) (10,0.27380090951919556) (11,0.2733047902584076) (12,0.2730652987957001) (13,0.2730197608470917) (14,0.27294835448265076) (15,0.2730795741081238) (16,0.27323007583618164) (17,0.2733716368675232) (18,0.27345335483551025) (19,0.273441880941391) (20,0.27346664667129517)};

\addplot[thick,solid,mark=triangle*,color=blue] coordinates {(0,0.27292895317077637)
(1,0.2738836705684662)
(2,0.2748337984085083)
(3,0.2755366563796997)
(4,0.2759058475494385)
(5,0.27563467621803284)
(6,0.2745468318462372)
(7,0.2723899483680725)
(8,0.2693990170955658)
(9,0.2657332420349121)
(10,0.26198145747184753)
(11,0.2584385871887207)
(12,0.255347341299057)
(13,0.2527577579021454)
(14,0.25060731172561646)
(15,0.2489776462316513)
(16,0.24770748615264893)
(17,0.2467743456363678)
(18,0.24587766826152802)
(19,0.24516364932060242)
(20,0.24469858407974243)};

\addplot[thick,solid,mark=square*,color=orange] coordinates {(0,0.2812603712081909)
(1,0.2822931706905365)
(2,0.2833845615386963)
(3,0.2842649221420288)
(4,0.2847609519958496)
(5,0.2846360504627228)
(6,0.28376084566116333)
(7,0.2819279134273529)
(8,0.27937325835227966)
(9,0.2762337625026703)
(10,0.27308934926986694)
(11,0.2700500786304474)
(12,0.2674236595630646)
(13,0.26527267694473267)
(14,0.26340344548225403)
(15,0.2620465159416199)
(16,0.2610079050064087)
(17,0.2602258324623108)
(18,0.2595546841621399)
(19,0.25897082686424255)
(20,0.258605033159256)};

\addplot[thick,solid,mark=diamond*,color=black] coordinates {(0,0.28910624980926514)
(1,0.29119643568992615)
(2,0.2935451567173004)
(3,0.2957402467727661)
(4,0.2976415753364563)
(5,0.2987489700317383)
(6,0.2988681495189667)
(7,0.29747480154037476)
(8,0.29481446743011475)
(9,0.29097622632980347)
(10,0.2866789698600769)
(11,0.28237053751945496)
(12,0.2783895432949066)
(13,0.2750270366668701)
(14,0.2720703184604645)
(15,0.2697802484035492)
(16,0.26798245310783386)
(17,0.2665613293647766)
(18,0.26527678966522217)
(19,0.26415562629699707)
(20,0.26336178183555603)};

\legend {Black Label, White Label, Multiracial Label, Biracial Label, Mixed Race Label};

\end{axis}
\end{tikzpicture}
\caption{Across 5\% increments in mixing ratio between Black source images (index 1) and White target images (index 21), the mean cosine similarity of Multiracial, Biracial, and Mixed Race labels varies with the mixing ratio of the images in a similar manner as the Black label, instead of increasing to a maximum value at index 11 (50\% mixing ratio), as would be expected if CLIP were detecting a mix of racial features. The Mixed Race label is preferred at the intermediate steps of the morph series.}
\label{mean_cos_sim_multiracial}
\Description{Figure depicting the mean cosine similarity at each step of the morph series with Black, White, and Multiracial labels.}
\end{figure*}

\begin{table*}[htbp]
{
\centering
\begin{tabular}
{|l||r|r|r|r|r|r|}
 \hline
 \multicolumn{7}{|c|}{Correlation (Pearson's $\rho$) of race or ethnicity label with "person" label (over $21,000$ images)} \\
 \hline
 Racial or Ethnic Group & A-W Female & B-W Female & L-W Female & A-W Male & B-W Male & L-W Male\\
 \hline
 Source Racial or Ethnic Group & \cellcolor{gray!13}0.13 & \cellcolor{gray!33}0.33 & \cellcolor{gray!21}0.21 & \cellcolor{gray!11}0.11 & \cellcolor{gray!40}0.40 & \cellcolor{gray!33}0.33 \\
 White & \cellcolor{gray!81}0.81 & \cellcolor{gray!75}0.75 & \cellcolor{gray!82}0.82 & \cellcolor{gray!60}0.60 & \cellcolor{gray!74}0.74 & \cellcolor{gray!64}0.64 \\
 \hline
\end{tabular}
}
\caption{Over $21,000$ images, the cosine similarity of the White label and the images strongly correlates with the cosine similarity of the person label and the images, with Pearson's $\rho$ up to $0.82$. Correlation does not exceed $0.40$ for any other label.}
\label{person_correlations_table}
\Description{Table describing the correlation of each race or ethnicity label with the "person" label.}
\end{table*}

\begin{table*}[htbp]
{
\centering
\begin{tabular}
{|l||r|r|r|r|r|r|}
 \hline
 \multicolumn{7}{|c|}{Standard Deviation of Cosine Similarities with $21,000$ Images by Morph Series} \\
 \hline
 Racial or Ethnic Group Label & A-W Female & B-W Female & L-W Female & A-W Male & B-W Male & L-W Male\\
 \hline
 Source Group & \cellcolor{gray!50}0.025 & \cellcolor{gray!38}0.019 & \cellcolor{gray!26}0.013 & \cellcolor{gray!56}0.028 & \cellcolor{gray!38}0.019 & \cellcolor{gray!28}0.014 \\
 White & \cellcolor{gray!16}0.008 & \cellcolor{gray!20}0.010 & \cellcolor{gray!16}0.008 & \cellcolor{gray!14}0.007 & \cellcolor{gray!16}0.008 & \cellcolor{gray!14}0.007 \\
 Person (Race Omitted) & \cellcolor{gray!12}0.006 & \cellcolor{gray!16}0.008 & \cellcolor{gray!12}0.006 & \cellcolor{gray!12}0.006 & \cellcolor{gray!12}0.006 & \cellcolor{gray!12}0.006 \\
 \hline
\end{tabular}
}
\caption{The standard deviation of associations between the White label and a series of $21,000$ images is smaller than for any other race or ethnicity. This indicates that the probability of the White label, like the person label, is nearly constant across the steps of the morph series, suggesting that White is a default race against which other races and ethnicities are defined.}
\label{std_dev_morph_series}
\Description{Table describing the standard deviation of associations with the Person, White, and Asian, Black, or Latino labels for each morph series.}
\end{table*}

\noindent\textbf{Evidence that White is the Default Racial Group in CLIP.} As shown in Table \ref{person_correlations_table}, the correlation between the cosine similarity of an image with person and cosine similarity of an image with White is higher in all morph series than the correlation between the cosine similarity of an image with person and with a minority racial group. Correlations are for $21,000$ images each, and $p$-values are $< 10^{-50}$ in all cases. While the mean cosine similarities with $1,000$ encoded images at each morph step vary according to mixing ratio for minority racial and ethnic groups, the mean cosine similarity of the White text label with encoded images is nearly invariant across the morph series, and nearly indistinguishable from the person text label. Table \ref{std_dev_morph_series} captures standard deviation by text label for $21,000$ images in all six morph series, and shows that standard deviations for the White label fall within $[0.007,0.010]$, while standard deviations for Black, Asian, and Latino/a labels fall within $[0.013,0.028]$, with all Black and Asian labels $\geq 0.019$. Standard deviations are small because cosine similarity ranges from 0 to 1, and has a narrow span for one kind of image (faces).

Adding Multiracial, Biracial, and Mixed Race labels reveals that, on average, the model associates intermediate morph steps with the Mixed Race text label rather than with Black or White, as shown in Figure \ref{mean_cos_sim_multiracial}. This result corresponds well to the research of \citet{peery2008black+}, who found evidence of hypodescent as an automatic, reflexive bias, but also found that humans acknowledge more complex racial identities when presented with option to choose them. However, cosine similarity with the Multiracial, Biracial, and Mixed Race text labels does not increase uniformly from both sides to a maximum at morph index 11 (50\% mixing ratio), as might be expected if the label denoted that the model has perceived a blend of features.  Rather, these labels have higher mean cosine similarity when the images are more similar to Black source images, and lower mean cosine similarity as images are more similar to White target images. That CLIP prefers the Mixed Race label of the three, and assigns lower probability to Multiracial than even the default White label, may be a consequence of the model being trained on internet image captions, such that more colloquial descriptions are preferred.

\begin{figure*}[htbp]
\begin{tikzpicture}
\begin{axis} [
    height=4cm,
    width=14cm,
    line width = .5pt,
    ymin = 0.35, 
    ymax = 0.70,
    xmin=-.5,
    xmax=20.5,
    ylabel={Mean Valence WEAT (Cohen's $d$)},
    ylabel shift=-5pt,
    xtick = {0,1,2,3,4,5,6,7,8,9,10,11,12,13,14,15,16,17,18,19,20},
    xticklabels = {1,2,3,4,5,6,7,8,9,10,11,12,13,14,15,16,17,18,19,20,21},
    xtick pos=left,
    ytick pos = left,
    title=Mean Unpleasant Valence Association by Morph Index - Black-White Male Series,
    xlabel= {Image Morph Index},
    legend style={at={(.50,.68)},anchor=south west}
]

\addplot[thick,solid,mark=*,color=red] coordinates {(0,0.6211256980895996) (1,0.6274209022521973) (2,0.6302261352539062) (3,0.6296887397766113) (4,0.6213808059692383) (5,0.6047797799110413) (6,0.5830597877502441) (7,0.555972158908844) (8,0.5259577035903931) (9,0.4914378821849823) (10,0.4638698399066925) (11,0.43957674503326416) (12,0.41975316405296326) (13,0.40845850110054016) (14,0.3987888693809509) (15,0.3968500792980194) (16,0.4009338617324829) (17,0.4063314199447632) (18,0.4157257080078125) (19,0.42736104130744934) (20,0.4397173225879669)};

\legend {Unpleasantness Bias in BM-WM Morph Series};

\end{axis}
\end{tikzpicture}
\caption{Images more similar to the Black Male source images (index 1) are more associated with unpleasantness than images similar to the White Male target images (index 21). Correlation between the mean association with the "Black" label and mean valence association across the morph series is Pearson's $\rho = .97$, $p < 10^{-12}$. This indicates that multiracial people are subject to valence bias in CLIP, and suggests a link between the model's perception of an image as "Black" and its automatic association with unpleasantness.}
\label{bias_magnitude_fig}
\Description{Figure depicting the mean unpleasantness association at each step of the Black to White Male morph series.}
\end{figure*}
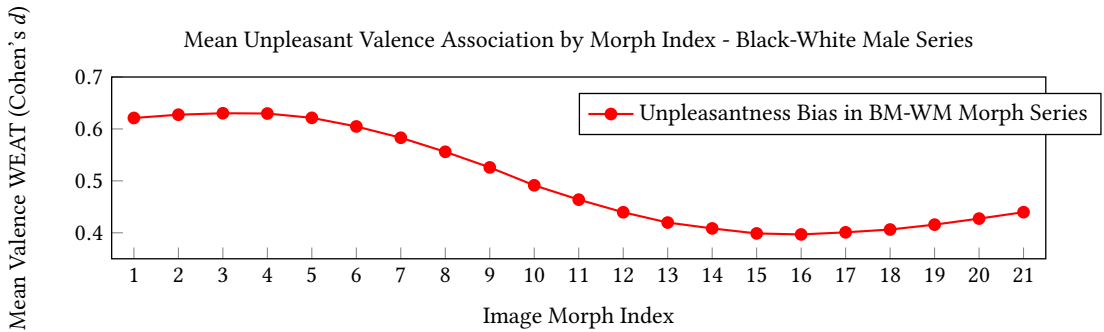

\noindent\textbf{Evidence that Valence Correlates with Minority Group Association.} As depicted in Figure \ref{bias_magnitude_fig}, mean valence association (association with bad or unpleasant vs. with good or pleasant) varies with the mixing ratio over the Black-White male morph series, such that CLIP encodes associations with unpleasantness for the faces most similar to CFD volunteers who self-identify as Black. As shown in Table \ref{valence_corr_table}, for the $21,000$ images of the Black-White male morph series, Pearson's $\rho$ between valence association and the cosine similarity of an image with the Black text label is $0.48$, $p < 10^{-90}$; $\rho$ between mean valence association and mean cosine similarity with the Black label at each morph step is $\rho = 0.97$, $p < 10^{-12}$. A similar effect is observed for the Black-White female morph series, with $\rho=0.41$, $p < 10^{-90}$ over $21,000$ images. The negative correlations for Asian-White morph series indicate that Asian faces are more associated with pleasantness relative to White faces. 

\begin{table}[htbp]
{
\centering
\begin{tabular}
{|l|r|r|r|r|}
\hline
\multicolumn{4}{|c|}{Correlation of Minority Group Label with Unpleasantness} \\
\hline
Female Series & Pearson's $\rho$ & Male Series & Pearson's $\rho$ \\
\hline
A-W Female & -0.22 ($10^{-90}$) & A-W Male & -0.43 ($10^{-90}$)\\
B-W Female & \cellcolor{gray!41}0.41 ($10^{-90}$) & B-W Male & \cellcolor{gray!48}0.48 ($10^{-90}$)\\
L-W Female & \cellcolor{gray!05}0.05 ($10^{-13}$) & L-W Male & \cellcolor{gray!01}0.01 (.26) \\
\hline
\end{tabular}
}
\caption{The valence association of an image correlates with the probability that it will be associated with a minority group text label. Most significantly, the unpleasantness association of an image increases with the probability that the model will associate the image with Black, with Pearson's $\rho = 0.48$, $p < 10^{-90}$ across $21,000$ images for the Black-White Male morph series.}
\label{valence_corr_table}
\Description{Table describing the correlation of unpleasantness associations with Asian, Black, or Latino associations, by morph series.}
\end{table}

\section{Discussion}

\textbf{The evidence indicates a rule of hypodescent, or one-drop rule, in CLIP.} The effect occurs in four of six morph series, and is more pronounced for images of women than images of men, such that the largest negative skew for male images is -0.06, for the Black-White male morph series, while the skew of associations for female images ranges between -0.30 (Black-White female series) and -0.84 (Asian-White female series). The three female morph series do not see a majority of associations with White until 40\% mixing ratio (Black-White series) or 20\% mixing ratio (Asian-White and Latina-White series). We describe a few possible causes for a primary effect based on gender. First, \citet{radford2021learning} observe that CLIP directs more attention to the physical features of women (such as hair), likely the result of training captions which describe women in terms of physical appearance. Such attention to physical features may render CLIP more sensitive to what the model perceives as race in women. On the other hand, \citet{halberstadt2011barack} have suggested that hypodescent in human subjects is related to attention, such that a perceiver notices deviation from the physical features most frequently observed. Prior research in computer vision and in NLP finds that women with darker skin \cite{buolamwini2018gender} or who are Black, Asian, or Latina \cite{wolfe2021low} are underrepresented in machine learning training data. Hypodescent as it pertains to women may thus be related to underrepresentation in CLIP's WIT training corpus.

\noindent\textbf{The evidence indicates that CLIP encodes White as a default race.} This is supported by the stronger correlations between White cosine similarities and person cosine similarities than for any other racial or ethnic group, with Pearson's $\rho \in [0.60, 0.82]$. Standard deviations for both person and White labels are also much lower than for Asian, Black, or Latino/a labels. This accords with the near invariance of the means of the White label across all steps of the morph series, such that images are associated with a minority racial or ethnic group based on deviation from White. That White is a default race in CLIP is further supported by image associations with Multiracial, Biracial, and Mixed Race text labels, which do not increase from both sides of the series to a peak at 50\% mixing ratio, but are higher when the image is similar to an image of a Black person, and lower when more similar to a White person.

\noindent\textbf{The evidence indicates that the valence of an image correlates with racial association}, with Pearson's $\rho = 0.48$, $p < 10^{-90}$ for images in the Black-White male series. More concretely, our results indicate that the more certain the model is that an image reflects a Black individual, the more associated with the unpleasant embedding space the image is. We note three consequences of this result. First, it indicates that multiracial people whom the model associates more strongly with Black are subject to valence biases in visual semantic AI. Second, it suggests that an implicit bias - of Black individuals with unpleasantness - is linked to the perception of race in visual semantic AI. This accords with research of \citet{zarate1990person}, who found that the speed of racial categorization by human perceivers is linked to the attribution of racial stereotypes to a subject. Third, this result affirms the results of smaller studies of implicit bias in word embeddings and in human subjects.

\noindent Observing a correlation between pleasantness and probability of the Asian text label may correspond to the "model minority" stereotype, wherein people of Asian ancestry are lauded for their upward mobility and assimilation into American culture, and even associated with "good behavior" \cite{wu2015color}.  This is likely a stereotype held in conjunction with Asian individuals being perceived and marked as outsiders \cite{lee2009model}, and the strong negative response during the pandemic \cite{tessler2020anxiety}. While this is consistent with CLIP's association of faces of Asian people relative to the White default group, further work is needed to confirm the underlying positive association to Asian and its dissociation from acts of discrimination.

\noindent\textbf{Impact on AI.} As indicated by the title of the paper introducing CLIP, the model is designed to learn "transferable" visual features, \textit{i.e.,} features which can be used in a wide variety of settings and applications \cite{radford2021learning}. One of these uses is to serve as ground truth for creating other specialized multimodal models: among the first uses of CLIP was to train the zero-shot image generation model DALL-E \cite{ramesh2021zero,radford2021learning}. A larger, non-public version of the CLIP architecture was used in the training of DALL-E 2 \cite{ramesh2022hierarchical}. Commensurate with the findings of the present research, the Risks and Limitations described in the DALL-E 2 model card note that it "produces images that tend to overrepresent people who are White-passing" \cite{ramesh2022hierarchical}. Such uses demonstrate the potential for the biases learned by CLIP to spread beyond the model's embedding space, as its features are used to guide the formation of semantics in other state-of-the-art AI models. Moreover, due in part to the advances realized by CLIP and similar models for associating images and text in the zero-shot setting, multimodal architectures have been described as the foundation for the future of widely used internet applications, including search engines \cite{nayak_2021}. Our results indicate that additional attention to what such models learn from natural language supervision is warranted.

\noindent\textbf{Impact on the Sciences.} Our results suggest that visual semantic AI models like CLIP may prove useful tools for studying societal-level biases. For example, word embeddings have been used to study the consistency of gender stereotypes across child and adult language corpora \cite{charlesworth2021gender}, as well as changes in gender and ethnic stereotypes over one hundred years \cite{garg2018word}. Visual semantic AI may facilitate similar scientific research enabling the study of human biases which can be observed in the statistical relationship between language and images.

\noindent\textbf{Impact on Society.} As human interaction with AI systems becomes more common, hypodescent and valence bias in visual semantic AI may affect the human perception of multiracial individuals. \citet{greenwald2015statistically} find that implicit racial biases in humans can "explain discriminatory impacts that are societally significant either because they can affect many people simultaneously or because they can repeatedly affect single persons." Given their ubiquity and ease of use, modern AI applications may, on an unprecedented scale, affect the way human beings perceive other human beings.

\noindent\textbf{Limitations and Future Work.} 
This research examines only one visual semantic model, which prevents assessment of whether our results generalize across architectures. CLIP is the first in a new generation of multimodal AI, and until late December 2021 was the only English-language zero-shot visual semantic model available open source for scientific study. Further work will be necessary to test hypodescent in new zero-shot visual semantic models such as SLIP \cite{mu2021slip}, and in other CLIP architectures. Moreover, CLIP creates highly contextual text representations, and using different labels may produce different results. We employ a principled method by adhering to the prompt outlined by \citet{radford2021learning}, and operationalizing only the racial and ethnic categories specified in the CFD. 2019 Google N-Gram frequency statistics indicate that these are also by far the most common ways to describe each racial or ethnic group in all English N-Gram sources \cite{lin2012syntactic}. Nonetheless, using different text prompts is likely to affect associations in the model, and might be explored in future work. Future work might also use a GAN-based approach such as that of \citet{lang2021explaining} to test what characteristics of an image directly influence hypodescent and valence bias. Such a study might test our hypothesis that increased attention to physical features results in stronger bias of hypodescent for women, a finding which varies from studies of hypodescent in human perceivers, for which hypodescent is observed more strongly for men \cite{ho2011evidence,sidanius2001social,navarrete2010prejudice}.

\section{Conclusion}

The primary result of the present research is the demonstration of hypodescent in a state-of-the-art visual semantic AI system. Additionally, the results reflect that women are more likely to be classified according to a rule of hypodescent, that White is a default race in the model, and that stereotype-congruent pleasantness bias correlates with association with Black. This work adds to a body of literature showing that AI encodes the implicit and explicit biases of the society and language on which it trains.

\begin{acks}
This material is based on research partially supported by the U.S. National Institute of Standards and Technology (NIST) Grant 60NANB\-20D212T. Any opinions, findings, and conclusions or recommendations expressed in this material are those of the authors and do not necessarily reflect those of NIST.
\end{acks}

\bibliographystyle{ACM-Reference-Format}
\bibliography{sample-base}

\appendix

\section{History of Multimodal AI}\label{sec:multimodal}
In designing CLIP, \citet{radford2021learning} build directly on the research of \citet{zhang2020contrastive}, who train ConVIRT, a medical image classifier which consists of an image encoder (ResNet 50) and a contextualizing language model (BERT \cite{devlin-etal-2019-bert}) trained on a contrastive learning objective. Foundational research on visual semantic embedding spaces was performed by \citet{socher2013zero}, who developed methods for zero-shot cross-modal transfer between image and language representations. \citet{frome2013devise} subsequently used a deep learning approach in DeViSE to embed images and text in a joint embedding space, allowing generalization to unseen classes. Other advances in vision-and-language AI have produced models such as VisualBERT, which aligns language and images and can ground relevant language to regions of an iamage \cite{li2019visualbert}, and VilBERT, which set new state-of-the-art on caption-based image retrieval \cite{lu2019vilbert}. More recent zero-shot image classifiers include the ALIGN model of \citet{jia2021scaling}, which trains on contrastive loss between a BERT language model and an EfficientNet-L2 \cite{xie2020self}, and the Turing-Bletchley model of \citet{tiwary_2021}, which realizes CLIP-like zero-shot capabilities for 94 languages. \citet{wang2021simvlm} introduce a visual semantic model trained on language modeling loss instead of contrastive loss, known as SimVLM. \citet{mu2021slip} introduce SLIP ("Self-Supervision meets Language-Image Pretraining"), a model which improves on CLIP's zero-shot results by adding view-based  self-supervision of images to the training objective.

\section{CLIP Download Statistics}

The number of downloads in the past month of the three CLIP models available on the Huggingface Transformers library is reported below, based on figures obtained from the library's website on April 26, 2022 \cite{wolf-etal-2020-transformers}:

\begin{itemize}
    \item CLIP ViT-Base-Patch32: 1,073,126 downloads
    \item CLIP ViT-Base-Patch16: 4,998 downloads
    \item CLIP ViT-Large-Patch14: 6,696 downloads
\end{itemize}

CLIP ViT-Base-Patch32 is also the default model used in example code provided in the official OpenAI Github repository for the model.

\section{Synthetic Image Series}\label{sec:synthetic}

\begin{figure*}
\begin{subfigure}
    \centering
    \includegraphics[width=14cm]{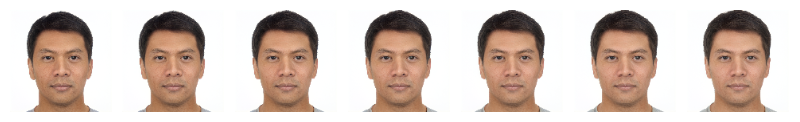}
\end{subfigure}

\begin{subfigure}
    \centering
    \includegraphics[width=14cm]{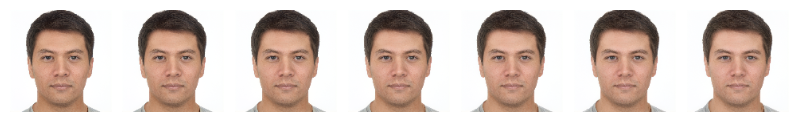}
\end{subfigure}

\begin{subfigure}
    \centering
    \includegraphics[width=14cm]{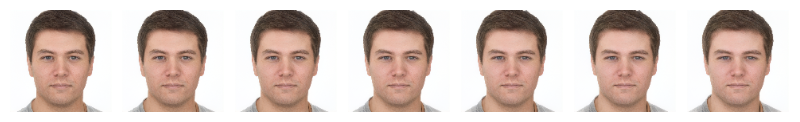}
\end{subfigure}
\caption{An Asian-White Male Morph Series}
\label{am_wm_full_series}
\Description{Example of Asian to White Male morph series.}
\end{figure*}

\begin{figure*}
\begin{subfigure}
    \centering
    \includegraphics[width=14cm]{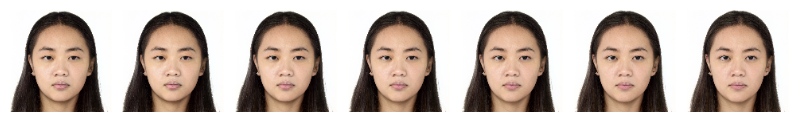}
\end{subfigure}

\begin{subfigure}
    \centering
    \includegraphics[width=14cm]{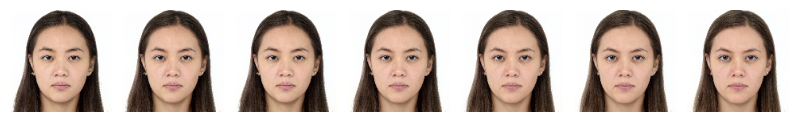}
\end{subfigure}

\begin{subfigure}
    \centering
    \includegraphics[width=14cm]{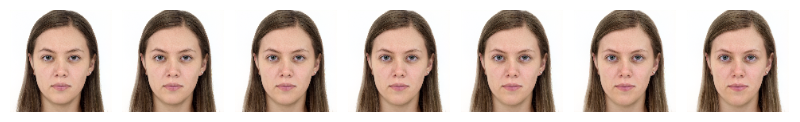}
\end{subfigure}
\caption{An Asian-White Female Morph Series}

\label{af_wf_full_series}
\Description{Example of Asian to White Female morph series.}

\end{figure*}

\begin{figure*}
\begin{subfigure}
    \centering
    \includegraphics[width=14cm]{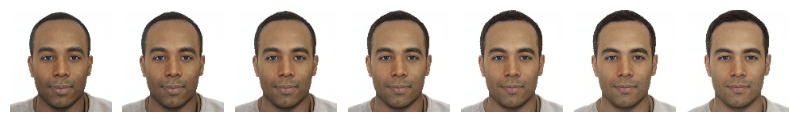}
\end{subfigure}

\begin{subfigure}
    \centering
    \includegraphics[width=14cm]{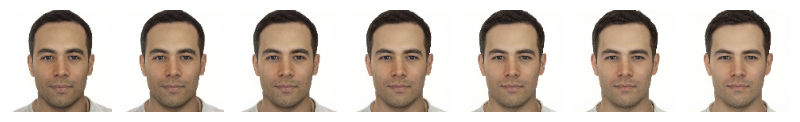}
\end{subfigure}

\begin{subfigure}
    \centering
    \includegraphics[width=14cm]{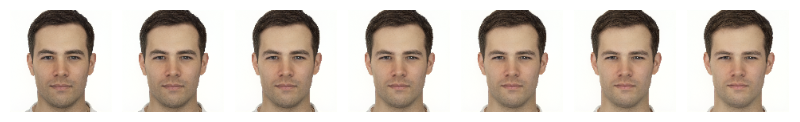}
\end{subfigure}
\caption{A Black-White Male Morph Series}

\label{bm_wm_full_series}
\Description{Example of Black to White Male morph series.}

\end{figure*}

\begin{figure*}
\begin{subfigure}
    \centering
    \includegraphics[width=14cm]{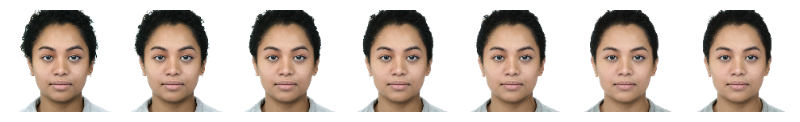}
\end{subfigure}

\begin{subfigure}
    \centering
    \includegraphics[width=14cm]{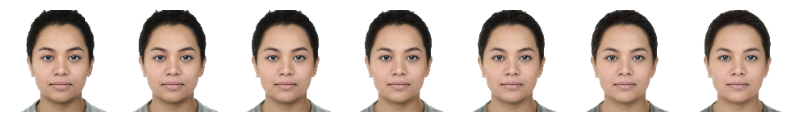}
\end{subfigure}

\begin{subfigure}
    \centering
    \includegraphics[width=14cm]{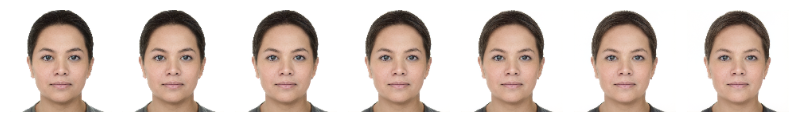}
\end{subfigure}
\caption{A Black-White Female Morph Series}
\label{bf_wf_full_series}
\Description{Example of Black to White Female morph series.}

\end{figure*}

\begin{figure*}
\begin{subfigure}
    \centering
    \includegraphics[width=14cm]{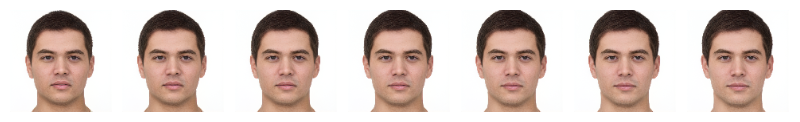}
\end{subfigure}

\begin{subfigure}
    \centering
    \includegraphics[width=14cm]{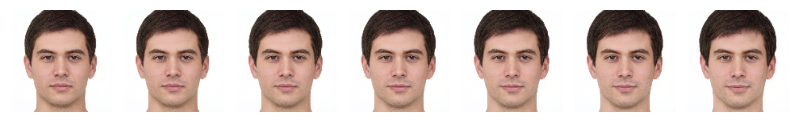}
\end{subfigure}

\begin{subfigure}
    \centering
    \includegraphics[width=14cm]{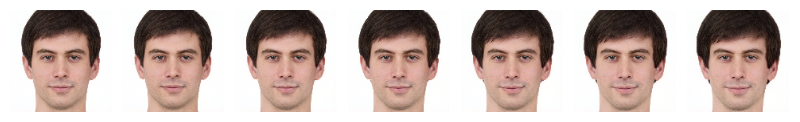}
\end{subfigure}
\caption{A Latino-White Male Morph Series}

\label{lm_wm_full_series}
\Description{Example of Latino to White Male morph series.}

\end{figure*}

\begin{figure*}
\begin{subfigure}
    \centering
    \includegraphics[width=14cm]{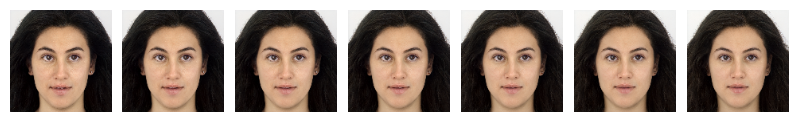}
\end{subfigure}

\begin{subfigure}
    \centering
    \includegraphics[width=14cm]{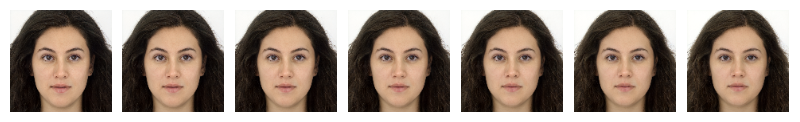}
\end{subfigure}

\begin{subfigure}
    \centering
    \includegraphics[width=14cm]{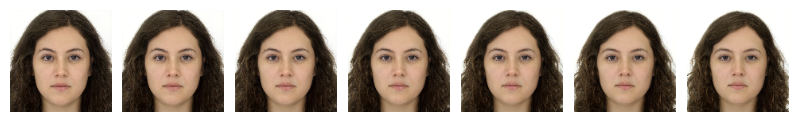}
\end{subfigure}
\caption{A Latina-White Female Morph Series}

\label{lf_wf_full_series}
\Description{Example of Latina to White Female morph series.}

\end{figure*}

To aid the reader in visualizing our experiments, we include via Figures \ref{am_wm_full_series} through \ref{lf_wf_full_series} below a single morph series for each of the six pairs of racial and ethnic groups we examine in this work. All images, including the first and last, are generated by StyleGAN2-ADA. Morph index 1 (source image of an Asian, Black, or Latino/a individual) corresponds to the upper left image, while morph index 21 (target image of an White individual) corresponds to the bottom right image. Note that the images output by the GAN are $1,024x1,024$ pixels, and the morph series we include are reduced in size to fit in this appendix but are fed to the CLIP preprocessor at full size.

\end{document}